\documentclass{article}
\usepackage{arxiv}

\usepackage[utf8]{inputenc} % allow utf-8 input
\usepackage[T1]{fontenc}    % use 8-bit T1 fonts
\usepackage{url}            % simple URL typesetting
\usepackage{booktabs}       % professional-quality tables
\usepackage{amsfonts}       % blackboard math symbols
\usepackage{nicefrac}       % compact symbols for 1/2, etc.
\usepackage{microtype}      % microtypography
\usepackage{lipsum}
\usepackage[table]{xcolor}
\usepackage{graphicx}
\usepackage{algorithm}      % For algorithm float environment
\usepackage{algpseudocode}   
\usepackage{multirow}
\usepackage{makecell} 
\usepackage{array}
\usepackage{tabularx}
\usepackage{subcaption}
\usepackage[skip=10pt]{caption} 
\usepackage{amsmath}        % For mathematical symbols and cases environment
\usepackage{amssymb}        % For additional math symbols
\usepackage{cite}
\usepackage{float}
\usepackage{colortbl}
\usepackage{hyperref}       % hyperlinks

\title{DPxFin: Adaptive Differential Privacy for Anti-Money Laundering
Detection via Reputation-Weighted Federated Learning}

\author{
 Renuga Kanagavelu \\
  A*STAR IHPC\\
  Singapore \\
  \texttt{renuga\_k@a-star.edu.sg} \\
  \And
 Manjil Nepal \\
  SRM University-AP\\
  Andhra Pradesh, India \\
  \texttt{manjil\_nepal@srmap.edu.in} \\
  \And
  Ning Peiyan \\
  EVYD\\
  Singapore \\
  \texttt{peiyan.ning@evydtech.com} \\
  \And
  Cai Kangning \\
  EVYD\\
  Singapore \\
  \texttt{kangning.cai@evydtech.com} \\
  \And
  Xu Jiming \\
  EVYD\\
  Singapore \\
  \texttt{jiming.xu@evydtech.com} \\
  \And
  Fei	Gao \\
  A*STAR IHPC\\
  Singapore \\
  \texttt{gaofei@a-star.edu.sg} \\
  \And
  Yong Liu \\
  A*STAR IHPC\\
  Singapore \\
  \texttt{liuyong@a-star.edu.sg} \\
  \And
  Goh Siow Mong Rick \\
  A*STAR IHPC\\
  Singapore \\
  \texttt{gohsm@a-star.edu.sg} \\
  \And
 Wei Qingsong \\
  A*STAR IHPC\\
  Singapore \\
  \texttt{wei\_qingsong@a-star.edu.sg} \\
}

\begin{document}
\maketitle
\begin{abstract}
In the modern financial system, combating money laundering is a critical challenge complicated by data privacy concerns and increasingly complex fraud transaction patterns. Although federated learning (FL) is a promising problem-solving approach as it allows institutions to train their models without sharing their data, it has the drawback of being prone to privacy leakage, specifically in tabular data forms like financial data. To address this, we propose \textbf{DPxFin}, a novel federated framework that integrates \textit{reputation-guided adaptive differential privacy}. Our approach computes client reputation by evaluating the alignment between locally trained models and the global model. Based on this reputation, we dynamically assign differential privacy noise to client updates, enhancing privacy while maintaining overall model utility. Clients with higher reputations receive lower noise to amplify their trustworthy contributions, while low-reputation clients are allocated stronger noise to mitigate risk. We validate \textbf{DPxFin} on the Anti-Money Laundering (AML) dataset under both IID and non-IID settings using Multi Layer Perceptron (MLP). Experimental analysis established that our approach has a more desirable trade-off between accuracy and privacy than those of traditional FL and fixed-noise Differential Privacy (DP) baselines, where performance improvements were consistent, even though on a modest scale. Moreover, \textbf{DPxFin} does withstand tabular data leakage attacks, proving its effectiveness under real-world financial conditions.
\end{abstract}


\newpage

%%------- Introduction  ----------%%
\section{Introduction}
Money laundering, involving the concealment of illicit funds from activities like drug trafficking, corruption, and tax evasion, poses serious risks to financial stability, economic growth, and national security. It is linked to major crimes such as terrorism funding and organised crime \cite{korejo2021concept}. The emergence of cryptoassets \cite{manning2025opportunities} has facilitated the increase of cross-border payments that are more anonymous, which has widened the risks of money laundering. The old methods of tracking are becoming obsolete, whereas AI and machine learning are promising to detect fraud better. Nevertheless, illicit flows are difficult to trace due to the complexity and anonymity of the new technologies as well as the tough legislative acts on data privacy \cite{beutel2020flower}. 

To address these challenges, financial institutions rely on traditional methods like AML regulations \cite{aidoo2025evaluating}, KYC protocols \cite{harrison2024ai}, and transaction monitoring systems \cite{zhang2024real}. 
Nevertheless, these methods are hindered by the changing crime strategies, the different regulatory environment, and the necessity to adjust to the new approach of the financial sector. Variation in regulation is a barrier to cross-border interaction, as every agency applies its regulations. Federated learning allows financial institutions to collaboratively train models without sharing data, enhancing fraud detection while preserving privacy \cite{narula2024comprehensive}. However, privacy leakage remains a concern, especially with sensitive data such as financial records. Studies, such as TabLeak \cite{vero2023tableak}, demonstrate that tabular data can be susceptible to attacks, hence emphasizing the requirement to have tight security solutions. Differential privacy helps protect against data leakage, but applying uniform noise across clients—who often have non-IID data—can harm model accuracy, making it challenging to balance privacy and performance.

To mitigate the problem of tabular data leakage during federated learning \cite{vero2023tableak}, we introduce \textbf{DPxFin}, a reputation-based dynamic differential privacy-based aggregation mechanism. Here the reputation of the clients is calculated at the server side using the Euclidean distance \cite{dokmanic2015euclidean} between each of the local models and a temporary aggregated global model. Clients of high reputation (clients whose updates best match the global model) are allocated low levels of differential privacy (DP) noise, resulting in the client contributions playing more of a larger effect on the global model. On the other hand, clients with a poor reputation are privileged with greater noise to protect model confidentiality and integrity. The reputation score is dynamically adjusted every round. \textbf{DPxFin} finds an optimal balance between the utility of the model and the privacy of the data. We evaluate the effectiveness of our proposed framework for financial fraud detection using the Anti-Money Laundering (AML) dataset \cite{altman2023realistic}, comparing it with various state-of-the-art methods. 

The key contributions of this work are summarized as follows, highlighting the primary innovations and methodological advancements introduced in this study:
\begin{enumerate}
    \item We introduce \textbf{DPxFin}, a novel reputation-driven differential privacy for leakage-resistant federated learning in finance.
    \item Our approach significantly enhances fraud detection performance by effectively handling non-IID data distributions, prioritising reliable client contributions.
    \item Extensive experiments on Anti-Money Laundering (AML) dataset demonstrate that our method achieves an optimal tradeoff between privacy and utility, outperforming existing approaches.
    \item We conduct a \textbf{TabLeak attack} \cite{vero2023tableak} on our federated learning system, validating its resilience and confirming the successful protection of financial data privacy.
\end{enumerate}

\section{Related Works}
Machine learning (ML) has become a critical tool in detecting money laundering, enhancing detection accuracy, scalability, and efficiency in the financial sector. This section summarises existing research and highlights key gaps in the field. A recent study \cite{guembe2023federated} examined the use of the federated learning model to identify money laundering with the  Federated Artificial Intelligence Enabler (FATE) framework and SecureBoost that preserves data privacy through homomorphic encryption. In the experimental setups, excellent accuracy and low false alarms of the centralised and federated architectures are observed. The centralised performance is more or less that of the federated model, with few differences. 

Similarly, \cite{nguyen2025privacy} proposed a privacy-preserving federated learning framework for anti-money laundering that integrates Random Forests with differential privacy and secure aggregation. Their model, evaluated on the AML dataset, achieved up to 85.71\% accuracy while addressing class imbalance and safeguarding data confidentiality using homomorphic encryption. Authors of the paper\cite{yu2024deep} studied how anti-money laundering (AML) can be strengthened using unsupervised learning patterns in a cross-border financial system. The work compares five DL architectures, such as CNN, hybrid CNN-GRU, etc. The best performances with regard to accuracy and area under the receiver operating characteristic curve (AUROC) were recorded with a self-developed hybrid Convolutional-Recurrent Neural Integration Model (CRNIM). 

In an attempt to address the shortcomings of the existing AML frameworks that do not incorporate the topological features and directional information contained in the transaction data, the authors of the paper \cite{liu2024federated} incorporated a bidirectional graph attention network (FALD-BGAT) deployed in a federated learning system. The clients train the model using the non-independent, homogeneously distributed data locally through the two-way graph attention network and send to a central server only the model parameters to be aggregated. The outcomes indicated high variation in the precision, recall, and F1-score with the already existing models. 

To mitigate the issue of a stand-alone strategy and lack of insights among financial institutions in detecting financial crimes, \cite{suzumura2022federated} put forward a new federated graph learning platform that enables joint training of models across financial institutions. Their model is a product of an association of federated learning and a graph-based modelling that learns fraud patterns that are complex and cross-institutional. Their contribution also focuses on the ability of collaborative intelligence to combat the monetarily focused crimes of the present internationally distributed crime system. A deep learning based anti-money laundering (AML) system was introduced by \cite{jensen2023qualifying} to leave behind the rule-based systems used before and find out the latent features of the sequence of transactions. Being more effective at qualifying and alarming, their best model employs gated recurrent units and self-attention layers. The model was put on a large data set of the Spar Nord Bank \cite{jensen2023synthetic} and thus proved to have monumental possibilities of identifying false positives of a conventional AML system and identifying missed high-risk clients exceptionally. A validation review made by an expert determined the utility of the flagged cases.

Authors \cite{alexandre2023incorporating} addressed the AML problem by designing a machine learning and risk assessment-based multiagent system to identify and report suspicious banking users. The system builds behavioural patterns using transactional data, legal rules, and expert AML knowledge to classify clients by risk and assist human investigators. It was tested on six months of real data with high accuracy. \cite{alkhalili2021investigation} proposed a machine learning-based solution to reduce false positives and inefficiencies in AML systems using a watch-list filtering mechanism. Their framework includes monitoring, advising, and action stages to automate transaction checks. Among the models tested, SVM with a polynomial kernel achieved the highest accuracy and was the first to automate watch-list filtering, a key compliance risk. Despite these advancements, most AML research still focuses on traditional machine learning and basic federated approaches, often overlooking user privacy and the risk of information leakage during training. Privacy-preserving techniques remain underdeveloped, limiting real-world deployment. For example, the differential privacy approach in \cite{guembe2023federated} lacks key practical details, such as privacy budgets or resilience to attacks. Similarly, \cite{yang2023anti} does not fully address privacy vulnerabilities, leaving systems open to inference risks. While \cite{nguyen2025privacy} demonstrates progress, our method surpasses it by a substantial margin in terms of accuracy, privacy preservation, and overall robustness. 

Differential Privacy (DP) protects client data by injecting noise into model updates; however, applying the same level of noise uniformly across all clients can degrade model performance, particularly in federated learning settings with non-IID data distributions. In such scenarios, some clients may provide more informative updates than others, and excessive noise can diminish the value of these contributions. Adaptive DP addresses this limitation by adjusting the noise applied to each client based on the significance of their updates, thereby achieving a better balance between privacy protection and model utility. To address these challenges, we propose a novel reputation-based adaptive differential privacy aggregation method, \textbf{DPxFin}, designed to effectively balance privacy guarantees and model performance in federated learning environments.

%%------- Methodology  ----------%%
\section{Methodology of DPxFin}
\begin{figure}[htbp]
    \centering
    \includegraphics[width=0.8\textwidth]{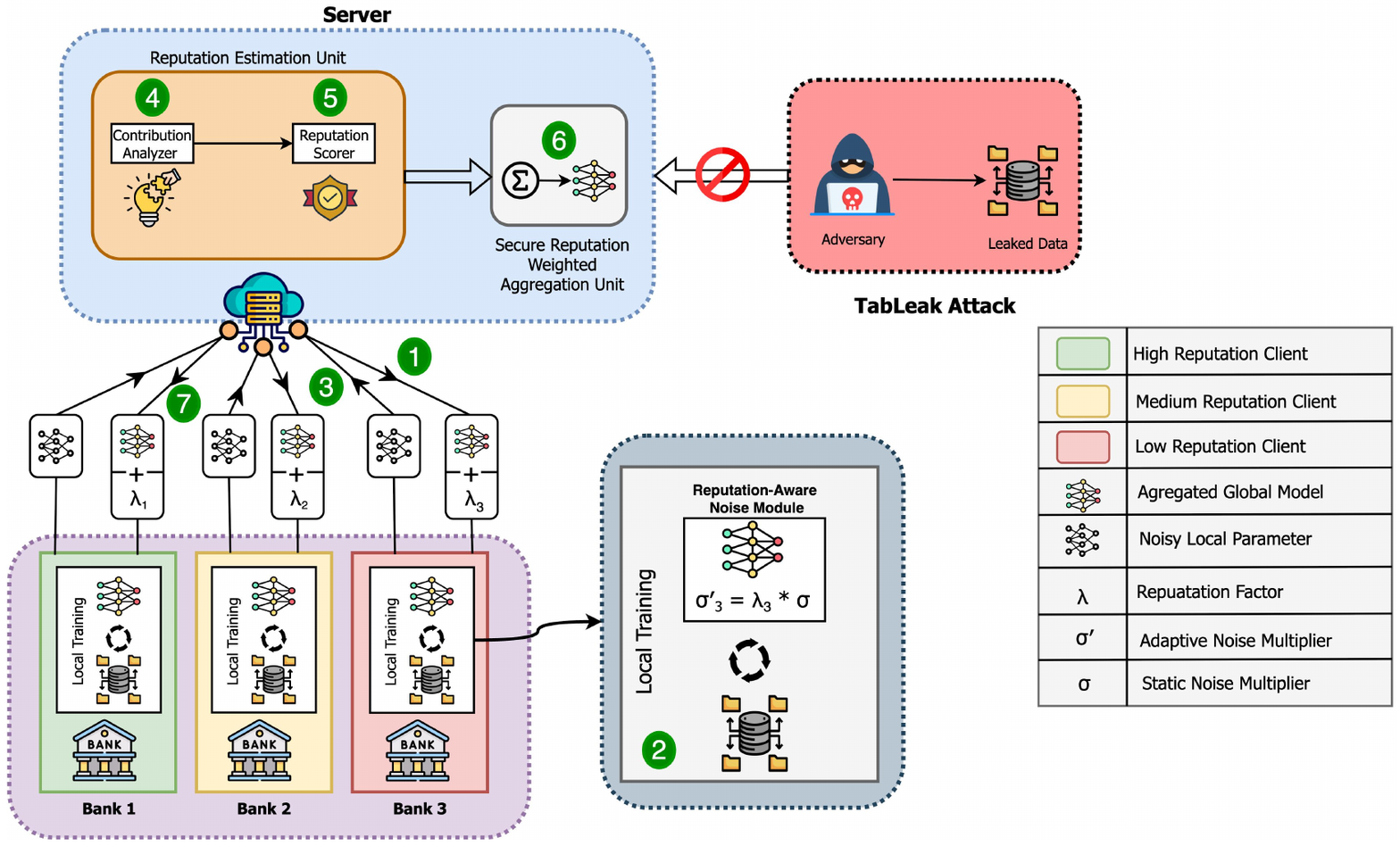}
    \caption{Framework Overview: (1) Server shares the global model and reputation factors (2) Clients perform local training using DP-SGD with adaptive noise (3) Noisy updates are returned (4) Server measures contributions (5) Calculates the reputation scores and updates reputation factors (6) Performs reputation weighted aggregation (7) Repeat the cycle.}
    \label{fig:proposed_arch}
\end{figure}

In this section, we describe our proposed \textbf{DPxFin} framework, illustrated in Figure~\ref{fig:proposed_arch}. The \textbf{DPxFin} framework enables secure and privacy-preserving collaborative model training by integrating adaptive differential privacy mechanisms into a federated learning setup coordinated by a central server. In this architecture, multiple distributed clients train models locally on their private data and share only privacy-protected updates with the central server for aggregation. By incorporating adaptive noise mechanisms and reputation-aware aggregation, \textbf{DPxFin} ensures that sensitive client information remains protected while still allowing distributed participants to collaboratively learn a robust and accurate global model. It comprises of three technical modules: \textbf{(1)} \textit{Feature Engineering and Data Preparation} which employs SMOTE (Synthetic Minority Over-sampling Technique) \cite{chawla2002smote} to synthetically balance minority class samples, mitigating class imbalance in money laundering detection; \textbf{(2)} \textit{Reputation-Based Adaptive Differential Privacy} integrates adaptive differential privacy by injecting reputation-based calibrated noise into each client’s local model updates, preserving data privacy while minimizing accuracy degradation; and \textbf{(3)} \textit{Server-Side Reputation Weighted Aggregation} performs a weighted aggregation of client updates, leveraging dynamic reputation scores to reduce susceptibility to model inversion or leakage attacks and to enhance the robustness and integrity of the global model. A high-level pseudocode of our proposed method, \textbf{DPxFin}, is provided in Algorithm~\ref{alg:dfxfin}, while the detailed components are described in Subsections~\ref{subsec:client_dp} and~\ref{subsec:server_agg}.

\begin{algorithm}[H]
\caption{DPxFin: Pseudo Code}
\label{alg:dfxfin}
\begin{algorithmic}[1]
\Require Global model $\mathcal{W}_0$, static noise multiplier $\sigma$, reputation factor $\lambda$, total rounds $T$, clients $\mathcal{C}$
\State Initialize reputation factor $\lambda = 1.0$ for all clients
\For{each round $t = 1$ to $T$}
    \State Select a subset of clients and share the global model
    \For{each selected client}
        \If{first round}
            \State Use static noise $\sigma$ as $\lambda$ is 1.0
        \Else
            \State Adjust noise using server-sent reputation factor
        \EndIf
        \State Train locally with noise and send model update
    \EndFor
    \State Aggregate updates into a temporary global model
    \State Compare each client’s update with the temporary model
    \State Calculate reputation scores and update reputation factors
    \State Re-aggregate updates weighted by reputation scores
\EndFor
\State \Return Final global model and updated reputation factors
\end{algorithmic}
\end{algorithm}

\subsection{Feature Engineering and Data Preparation} \label{data_preprocess}
We utilized the Anti-Money Laundering (AML) dataset \cite{altman2023realistic}, which simulates financial transactions—such as bank transfers, purchases, credit card payments, and checks—generated within a multi-agent virtual environment. This dataset is characterised by a severe class imbalance, with fraudulent transactions representing less than 1\% of the total records. Such an imbalance often biases models toward the majority legitimate class, reducing their effectiveness in detecting illicit activities. To address this, we applied the SMOTE technique \cite{chawla2002smote} to oversample the minority (money laundering) class, reducing class imbalance and expanding the dataset to over 7.5 million records for model training. 

From the timestamp column, we derived additional temporal features such as the transaction hour, day of the week, and month to capture time-based behavioural patterns associated with fraudulent activities. This feature engineering step enriched the dataset with meaningful temporal context, potentially improving the model’s ability to distinguish subtle patterns in illicit behaviour. Out of the 7.5 million records, 80\% (approximately 6 million) were used for training and distributed among the clients. The remaining 20\% (approximately 1.5 million) was reserved as a held-out test set to evaluate the performance of the best global model using classification metrics after training was completed.

\subsection{Reputation-Based Adaptive Differential Privacy}\label{subsec:client_dp}
Each client (local bank) \(k\) holds a local dataset $\mathcal{D}_k$ and computes model updates \(\Delta w_k\) based on its local training. To preserve privacy, each client clips its update to a fixed norm bound \(C\) to limit sensitivity:

\begin{equation}
\bar{\Delta w}_k = \frac{\Delta w_k}{\max\left(1, \frac{\|\Delta w_k\|_2}{C}\right)}
\end{equation}

In the initial training round, where no reputation scores exist, the client adds Gaussian noise using a fixed noise multiplier \(\sigma\) to ensure differential privacy:

\begin{equation}
\tilde{\Delta w}_k = \bar{\Delta w}_k + \mathcal{N}(0, (\sigma C)^2 I)
\end{equation}
\\
This noisy model update \(\tilde{\Delta w}_k\) is then sent to the server. Upon receiving updates from all participating clients, the server first computes a reputation score based on the Euclidean distance between each client’s update and a temporary global model as defined in Equation~\eqref{eq:euclidean_eq}. Using this score, a reputation factor \(\lambda_k \in (0, 1)\) is then calculated for each client. In subsequent rounds, the server communicates the reputation factor \(\lambda_k\) back to each client, which is then used to adaptively scale the noise during local training.

\begin{equation}
\sigma_k' = \lambda_k \cdot \sigma
\end{equation}
\\
Accordingly, the client adds Gaussian noise with this adjusted noise multiplier \(\sigma_k'\):

\begin{equation}
\tilde{\Delta w}_k = \bar{\Delta w}_k + \mathcal{N}(0, (\sigma_k' C)^2 I)
\end{equation}

This kind of reputation based-differential privacy scheme allows clients with better reputation scores (i.e clients that have regularly provided good and trustful updates) to add a relatively small noise to their model updates, which has a greater utility and convergence. Conversely, more noise is assigned to clients with lower reputations on a proportional basis which improves their privacy assurances and makes the system more resistant to unreliable or maliciously malicious updates. Compared to the traditional fixed-noise strategies that treat all clients with uniform noise without regard to their behavior, this adaptive-noise protocol provides greater privacy-utility trade-off.

The client-side training pseudocode is given in Algorithm~\ref{alg:client-side-training}.

\begin{algorithm}[H]
{\fontsize{10}{11}\selectfont
\caption{Client-Side Training}
\label{alg:client-side-training}
\begin{algorithmic}[1]
\State \textbf{Input:} Local dataset $\mathcal{D}_k$, global model $\mathcal{W}_t$, reputation factor $\lambda_k$, static noise multiplier $\sigma$, server round $t$
\If{$t = 1$}
    \State Set dynamic noise multiplier: $\sigma_k' = \sigma$
\Else
    \State Compute dynamic noise multiplier: $\sigma_k' = \lambda_k \cdot \sigma$
\EndIf
\State Train model using DP-SGD with noise multiplier $\sigma_k'$ on $\mathcal{D}_k$ to obtain local model $\tilde{\Delta w}_k$
\State Return $\tilde{\Delta w}_k$ to server
\end{algorithmic}
}
\end{algorithm}

%----------------------------------------------------------------
\subsection{Server-Side Reputation Weighted Aggregation Phase}\label{subsec:server_agg}
Following the receipt of the differentially private local updates \(\tilde{\Delta w}_i\) from participating clients, the server performs two key operations: (1) Computation of each client's reputation score $\text{rep}_i$ and reputation factor $\lambda_i$ (2) Aggregation of local models weighted by these reputation scores $\text{rep}_i$. To assess each client's alignment with the global objective, the server first computes the Euclidean distance \(d_i\) between each client's model update and a preliminary global aggregation \(\mathcal{W}_{\text{agg}}\), obtained via standard averaging:

\begin{equation}
d_i = \left\| \tilde{w}_i - \mathcal{W}_{\text{agg}} \right\|_2
\label{eq:euclidean_eq}
\end{equation} 
\\
Smaller distances reflect closer alignment with the global model, thus indicating higher quality contributions. To ensure that lower distances yield higher scores, a normalized consistency score is computed as:
\begin{equation}
\tilde{d}_i = 1 - \frac{d_i}{\max_j d_j}
\end{equation}
Now the raw reputation score is calculated as:
\begin{equation}
\text{rep}_i = \tilde{d}_i 
\end{equation}
 The raw reputation scores are then normalised across all \(N\) clients:
\begin{equation}
\bar{\text{rep}}_i = \frac{\text{rep}_i}{\sum_{j=1}^{N} \text{rep}_j}
\end{equation}

Next, the reputation factor \(\lambda_i\) for each client is assigned using a tiered strategy based on their individual current normalised reputation score:

\begin{equation}
\lambda_i = 
\begin{cases}
0.2, & \text{if } \bar{\text{rep}}_i \geq P_{70} \\
0.5, & \text{if } P_{50} \leq \bar{\text{rep}}_i < P_{70} \\
1.0, & \text{otherwise}
\end{cases}
\end{equation}

where \(P_{70}\) and \(P_{50}\) represent the 70th and 50th percentiles of all clients' current normalised reputation scores, respectively, and $\bar{\text{rep}}_i$ is the current normalised reputation score of client \(i\). \\

This dynamic adjustment ensures that clients making strong contributions to the global model are penalized less in the differential privacy mechanism (via lower noise), while weaker or potentially untrustworthy clients are subjected to higher privacy-preserving noise. This facilitates faster convergence and enhances the overall robustness of the model. Finally, the server aggregates the local models using the normalised reputation scores:
\begin{equation}
\mathcal{W}_{t+1} = \sum_{i=1}^{N} \bar{\text{rep}}_i \cdot \tilde{w}_i
\end{equation}
\\
The updated global model \(\mathcal{W}_{t+1}\) and the corresponding set of reputation factors \(\{\lambda_i\}_{i=1}^{N}\) are then distributed to the respective clients for the next training round.

\section{Experimental Setup}
In this section, we outline the experimental setup needed to implement the proposed model, evaluate its performance, and assess its effectiveness.

\subsection{Dataset}
\begin{figure}[h]
    \centering
    \includegraphics[width=0.65\textwidth]{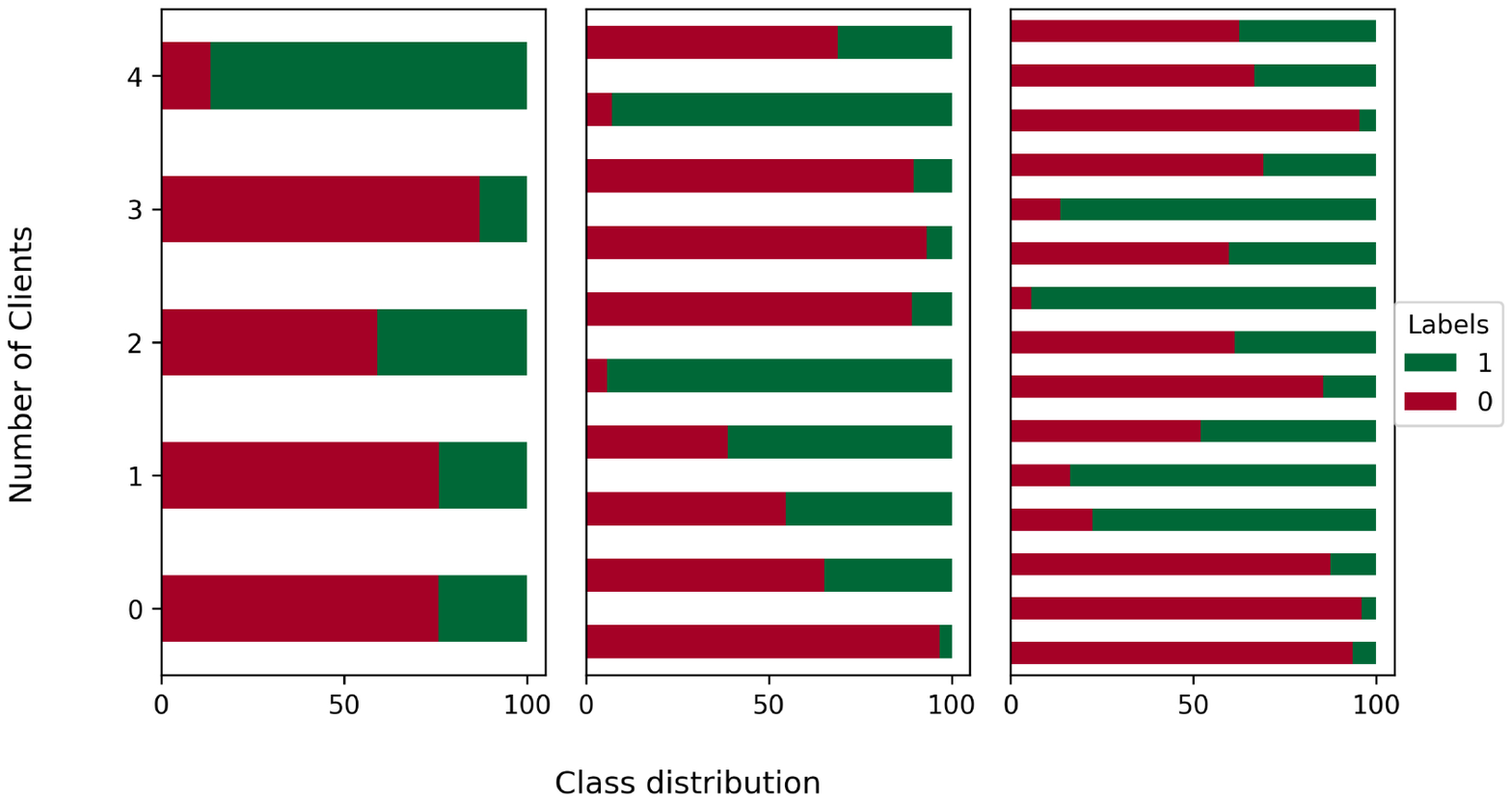}
    \caption{Dataset distribution among clients under a non-IID setting.}
    \label{fig:non_iid_distribution_drich}
\end{figure}

The IBM Synthetic Financial Data Money Laundering dataset \cite{altman2023realistic} is used in our experiments, specifically the HI\_Small Transaction dataset, which contains approximately 5 million rows, with around 5,000 positive examples of money laundering transactions. The dataset includes features such as transaction amount, timestamp, sender and receiver IDs, and transaction types. It is synthesised to mimic real-world transactions, allowing for controlled testing of machine learning models. The data, preprocessed as described in the Subsection~\ref{data_preprocess}, was divided into IID and non-IID settings; for IID, the data was randomly split equally among clients, while for non-IID, partitioning was based on a Dirichlet distribution \cite{yurochkin2019bayesian} with a concentration parameter of $\alpha = 1.0$, creating a moderately biased dataset with some class overlap as shown in Fig~\ref{fig:non_iid_distribution_drich}. 

Also, Table~\ref{tab:small_ibm_features} presents the feature schema of the financial transaction dataset used in this study. The table provides an overview of each feature, its corresponding data type, and its role in representing different aspects of financial transactions used for model training and evaluation.

\begin{table}[h]
\centering
\fontsize{10}{11}\selectfont
\renewcommand{\arraystretch}{1.2}
\begin{tabular}{|p{2.8cm}|p{1.2cm}|p{4cm}|}
\hline
\textbf{Feature} & \textbf{Type} & \textbf{Description} \\
\hline
From Bank & int64 & Source bank ID \\
\hline
To Bank & int64 & Destination bank ID \\
\hline
Amount Received & float64 & Received amount \\
\hline
Receiving Currency & int64 & Currency code (received) \\
\hline
Amount Paid & float64 & Paid amount \\
\hline
Payment Currency & int64 & Currency code (payment) \\
\hline
Payment Format & int64 & Payment method ID \\
\hline
Year & int64 & Transaction year \\
\hline
Month & int64 & Transaction month \\
\hline
Day & int64 & Transaction day \\
\hline
Hour & int64 & Transaction hour \\
\hline
Is Laundering & int64 & Binary label (0/1) \\
\hline
\end{tabular}
\vspace{2mm}
\caption{Feature Schema of AML Dataset}
\label{tab:small_ibm_features}
\end{table}

\subsection{Computational Resources and Frameworks}
Our experiments were conducted using a high-performance computing environment, summarized in Table~\ref{tab:config}. The experimental setup was designed to support efficient training and evaluation of federated learning models across multiple clients while maintaining reproducibility and scalability. The software stack consists of Python as the primary programming environment, along with PyTorch \cite{paszke2019pytorch} for deep learning model development and training. In addition, we utilized the Flower federated learning framework \cite{beutel2020flower}, which provides a flexible and scalable infrastructure for simulating federated learning workflows and coordinating communication between distributed clients and the central server.

To ensure privacy-preserving model training, we incorporated differential privacy (DP) mechanisms using the Opacus library \cite{yousefpour2021opacus}. Opacus is specifically designed to integrate seamlessly with PyTorch-based models and provides tools for implementing gradient clipping and noise injection to enforce formal differential privacy guarantees during training. By leveraging these tools, we were able to simulate realistic federated learning scenarios while protecting sensitive client-level information contained in local datasets. This setup enabled systematic experimentation with privacy-preserving federated learning while maintaining an effective balance between model utility and privacy protection.

\begin{table}[H]
\centering
\renewcommand{\arraystretch}{1.55}
\small
\begin{tabular}{|l|l|}
\hline
\textbf{Component} & \textbf{Specification} \\
\hline
\multicolumn{2}{|l|}{\textbf{\textit{Software Environment}}} \\
\hline
Programming Language & Python 3.11 \\
\hline
Deep Learning Framework & PyTorch 2.6.0+cu124 \\
\hline
Federated Learning Framework & Flower 1.9.0 \\
\hline
Differential Privacy Library & Opacus\\
\hline
\multicolumn{2}{|l|}{\textbf{\textit{Hardware Platform}}} \\
\hline
GPU & NVIDIA DGX with V100 GPUs \\
\hline
\end{tabular}
\vspace{2mm}
\caption{Experiment Setup}
\label{tab:config}
\end{table}

\section{Evaluation of Performance Metrics}

To evaluate model performance, we computed standard classification metrics using the predicted and true labels. These metrics are based on the counts of True Positives (TP), False Positives (FP), True Negatives (TN), and False Negatives (FN), which are fundamental to understanding how the model performs across different classes.

\begin{itemize}
    \item \textbf{Accuracy}: It is the proportion of correctly predicted instances (both positive and negative) out of all predictions, reflecting the model’s overall correctness. 
    \begin{equation}
    \text{Accuracy} = \frac{TP + TN}{TP + TN + FP + FN}
    \end{equation}

    \item \textbf{Recall}: It measures the model’s ability to correctly identify all relevant instances of a particular class.
    \begin{equation}
    \text{Recall} = \frac{TP}{TP + FN}
    \end{equation}

    \item \textbf{Precision}: It quantifies how many of the predicted positive cases are actually correct.
    \begin{equation}
    \text{Precision} = \frac{TP}{TP + FP}
    \end{equation}
    
    \item \textbf{F1 Score}: It is the harmonic mean of precision and recall, balancing false positives and false negatives in imbalanced datasets.
    \begin{equation}
    \text{F1-score} = 2 \times \frac{\text{Precision} \times \text{Recall}}{\text{Precision} + \text{Recall}}
    \end{equation}
\end{itemize}

\section{Result and Analysis}
In this section, we present the experimental results to evaluate the performance of the proposed \textbf{DPxFin} framework. We analyze its effectiveness under different privacy budgets and assess its resilience against privacy attacks such as TabLeak. The results highlight the trade-offs between privacy and utility and demonstrate the robustness of our approach in both IID and non-IID federated learning scenarios.

% \subsection{Comparison with various state-of-art (SOTA) methods}
% We present a comparative analysis of the proposed method with state-of-the-art differential privacy techniques on the AML dataset.

\subsection{Analysis Under Different Privacy Budget}
\begin{figure}[H]
\centering
\begin{subfigure}{0.49\linewidth}
    \centering
    \includegraphics[width=\linewidth]{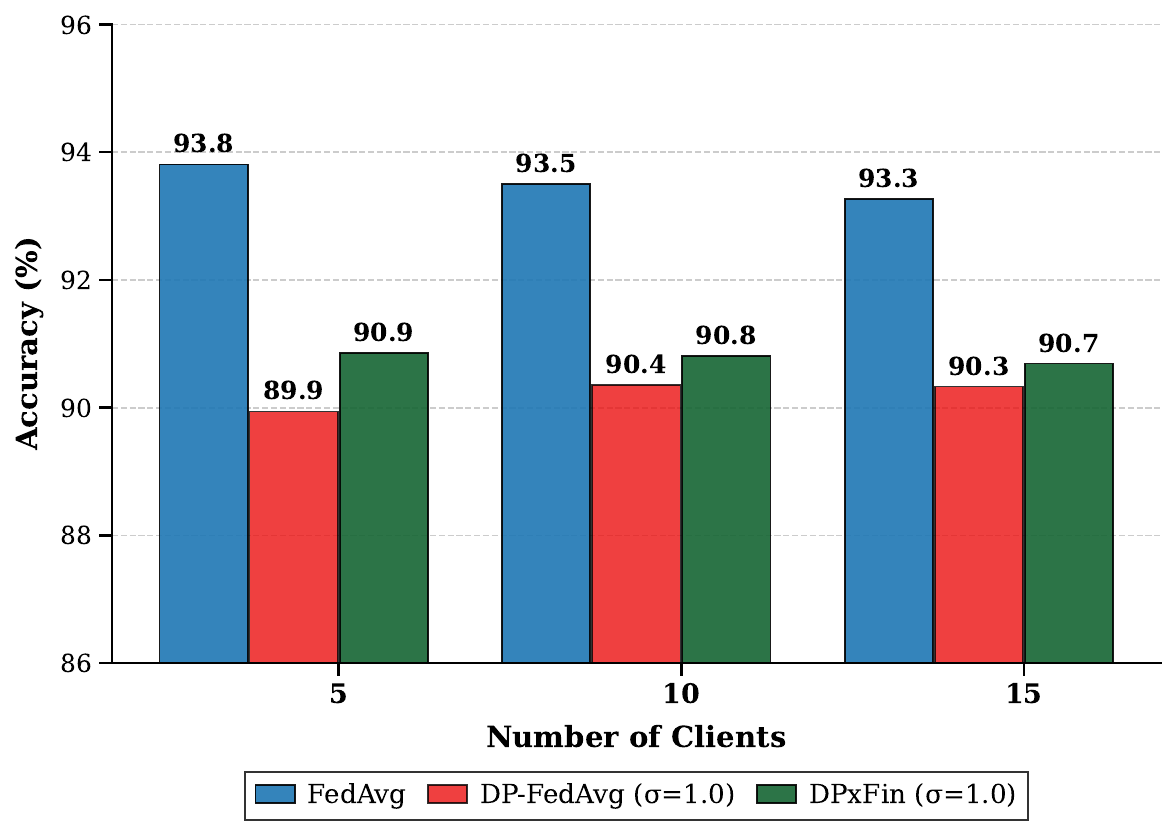}
    \caption{IID Distributions}
    \label{fig:barplot-iid}
\end{subfigure}
\hfill
\begin{subfigure}{0.49\linewidth}
    \centering
    \includegraphics[width=\linewidth]{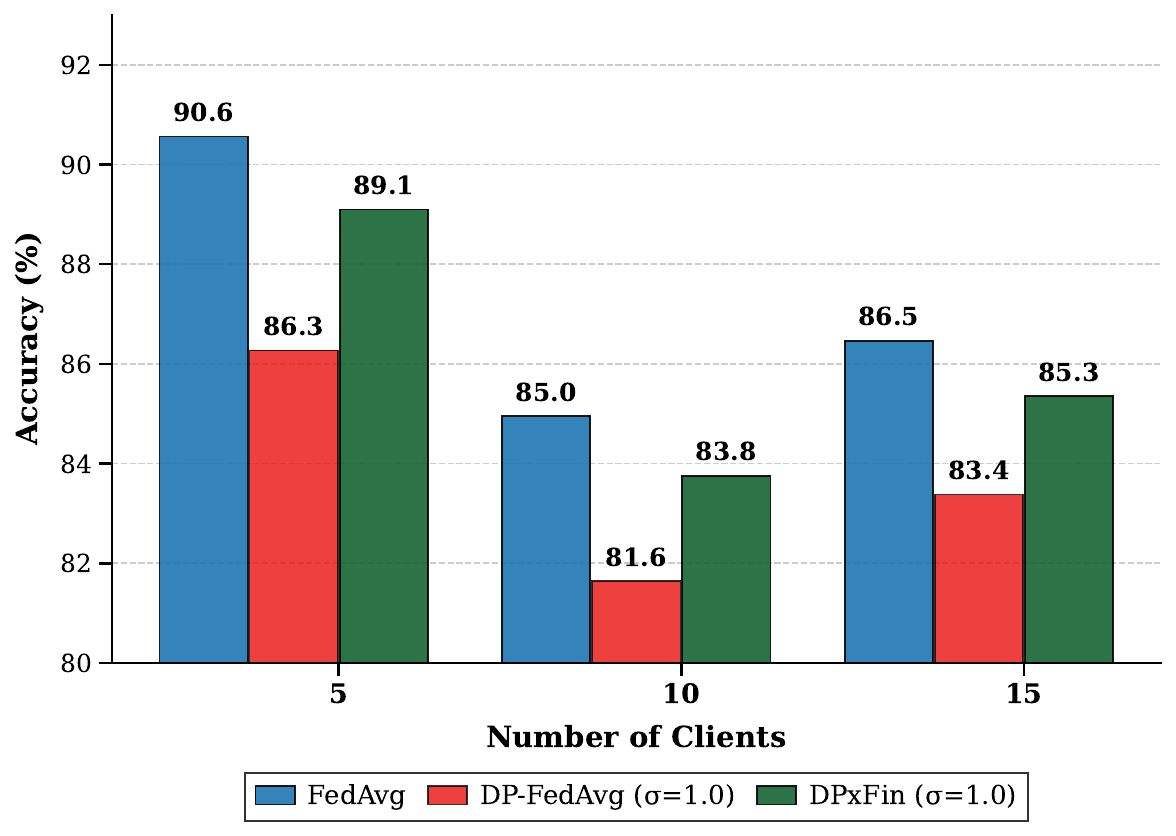}
    \caption{Non-IID Distributions}
    \label{fig:barplot-non-iid}
\end{subfigure}

\caption{Accuracy comparison under IID and Non-IID distributions}
\label{fig:barplot-combined}
\end{figure}

To evaluate the effectiveness of our proposed \textbf{DPxFin} method, we compare its accuracy with federated learning without privacy (FedAvg), and federated training with fixed privacy (DP-FedAvg). Figure~\ref{fig:barplot-iid} and ~\ref{fig:barplot-non-iid} show the accuracy for IID and Non-IID settings computed at the server after each training round. Each client evaluates the current global model on its local test dataset and sends the resulting local accuracy, along with the number of test samples they have to the server. The server then calculates the accuracy by multiplying each client's local accuracy with its number of test samples and dividing the total by the sum of all test samples. The comparison is conducted for both IID and non-IID settings on Multi Layer Perceptron (MLP). Our proposed method improves accuracy while effectively balancing the tradeoff between privacy and utility.

Compared to federated training with fixed privacy (DP-FedAvg), our proposed method consistently outperformed in terms of accuracy, achieving an max improvement of around 3\% in non-IID client variations. These pronounced gains in the non-IID scenario highlight the robustness and practical applicability of our approach in real-world, heterogeneous environments.

\begin{figure*}[t]
    \centering
    % First row with 3 subfigures
    \begin{subfigure}[t]{0.33\textwidth}
        \centering
        \includegraphics[width=1.05\linewidth]{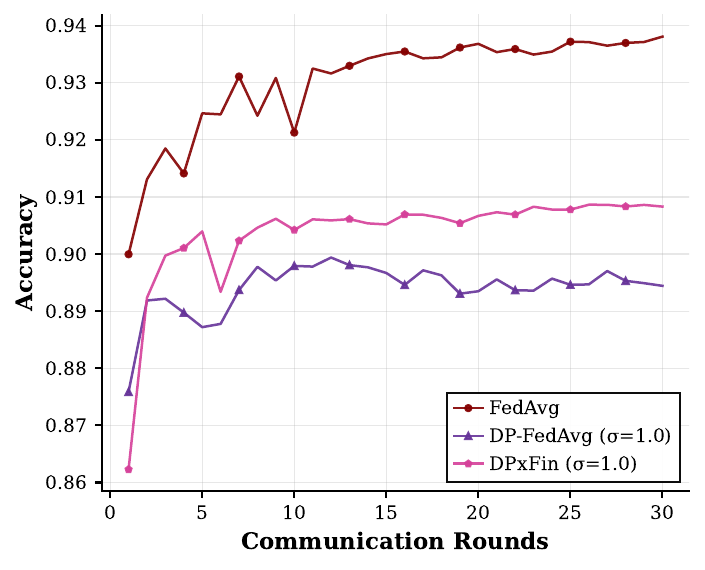}
        \caption{Test Accuracy For 5 Clients (IID)}
        \label{fig:iid-five}
    \end{subfigure}
    \hfill
    \begin{subfigure}[t]{0.33\textwidth}
        \centering
        \includegraphics[width=1.05\linewidth]{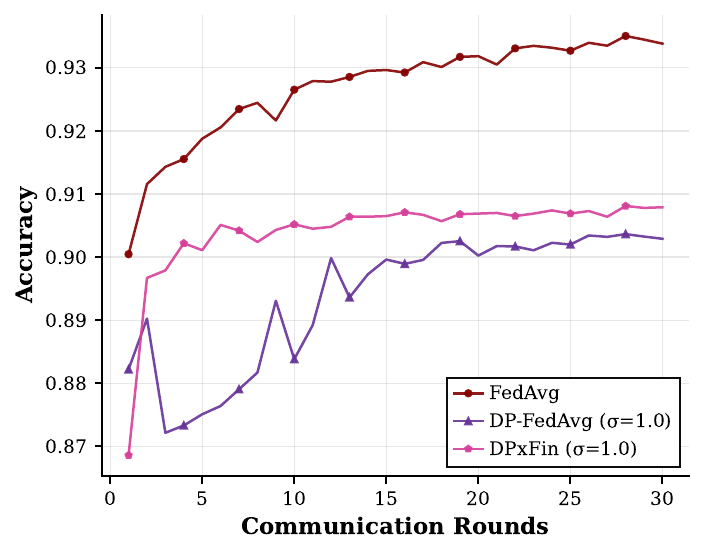}
        \caption{Test Accuracy For 10 Clients (IID)}
        \label{fig:non-iid-five}
    \end{subfigure}
    \hfill
    \begin{subfigure}[t]{0.33\textwidth}
        \centering
        \includegraphics[width=1.05\linewidth]{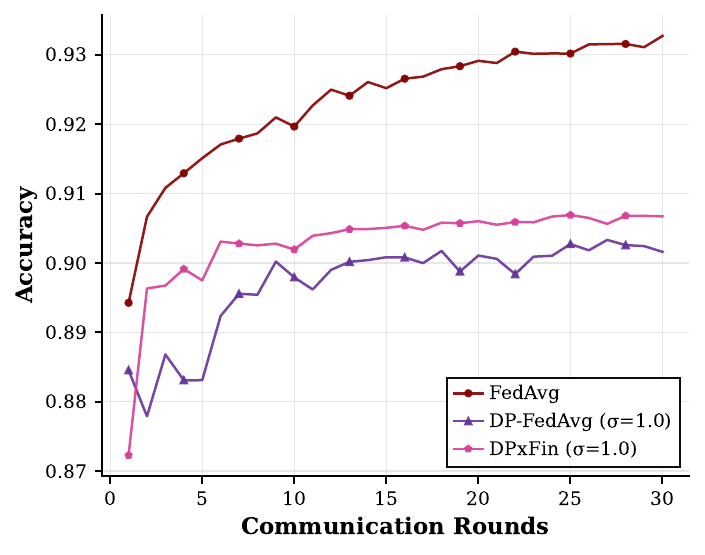}
        \caption{Test Accuracy For 15 Clients (IID)}
        \label{fig:iid-ten}
    \end{subfigure}
    
    \vspace{0.5cm}
    
    % Second row with 3 subfigures
    \begin{subfigure}[t]{0.33\textwidth}
        \centering
        \includegraphics[width=1.05\linewidth]{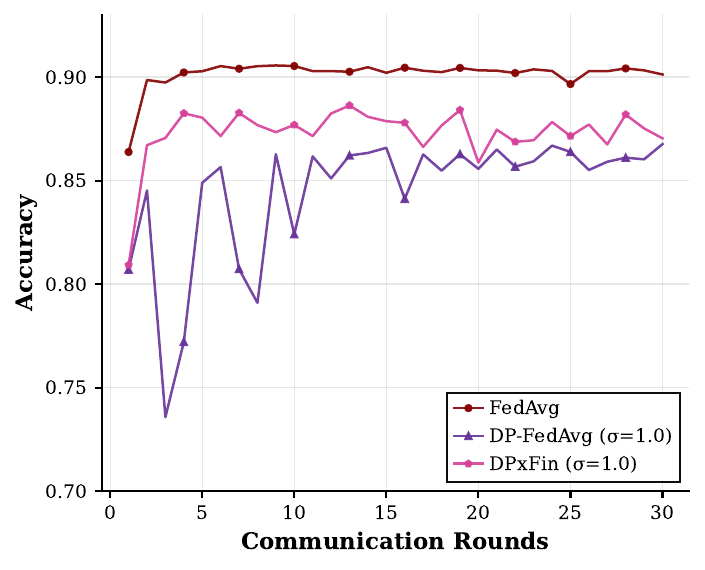}
        \caption{Test Accuracy For 5 Clients (Non-IID)}
        \label{fig:non-iid-ten}
    \end{subfigure}
    \hfill
    \begin{subfigure}[t]{0.33\textwidth}
        \centering
        \includegraphics[width=1.05\linewidth]{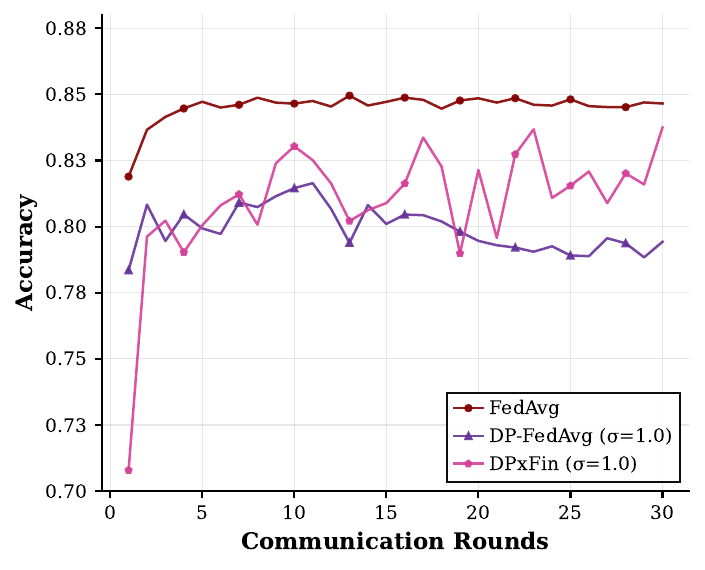}
        \caption{Test Accuracy For 10 Clients (Non-IID)}
        \label{fig:iid-fifteen}
    \end{subfigure}
    \hfill
    \begin{subfigure}[t]{0.33\textwidth}
        \centering
        \includegraphics[width=1.05\linewidth]{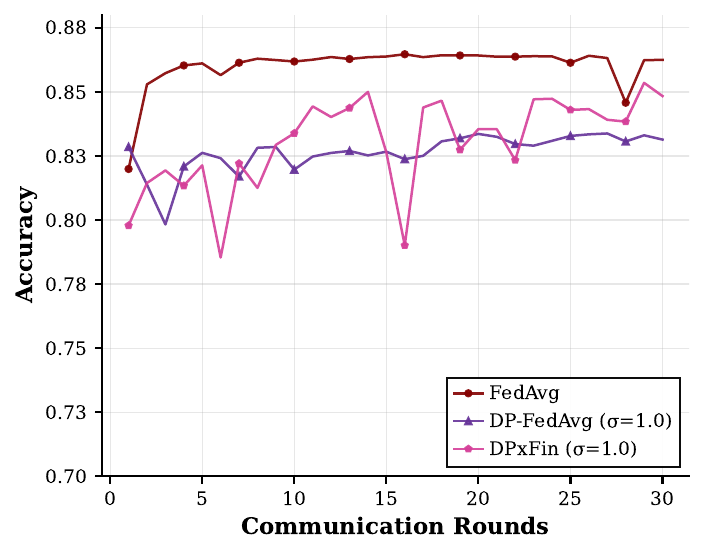}
        \caption{Test Accuracy For 15 Clients (Non-IID)}
        \label{fig:non-iid-fifteen}
    \end{subfigure}
    
    \caption{Performance Across Different Federated Settings}
    \label{fig:lineplot-comparison}
\end{figure*}

Figure~\ref{fig:lineplot-comparison} presents the model’s accuracy across communication rounds under both IID and non-IID data settings. This comparison highlights how different data distributions impact the model’s learning behavior and convergence over time.

\subsection{Tabular Attack}
To evaluate the risk of data leakage in federated learning on tabular datasets, we employed the TabLeak \cite{vero2023tableak} attack. Table~\ref{tab:tableak_results} reports the accuracy and standard deviation for both the baseline without noise (FedAvg) and our proposed DPxFin method. The high accuracy of 92.9\% for the baseline confirms that TabLeak can successfully reconstruct sensitive information, highlighting the vulnerability of standard federated learning pipelines. In contrast, DPxFin, which introduces reputation-sensitive noise, reduces the attack accuracy to 58.5\%, demonstrating its effectiveness in mitigating data leakage. These results underscore the efficacy and robustness of our noise-scaling approach, ensuring privacy preservation without compromising fairness, making it well-suited for real-world federated learning applications involving sensitive data.

% \begin{table}[h]
% \renewcommand{\arraystretch}{1.5}
% \centering
% \small
% \begin{tabular}{|c|c|c|c|}
% \hline
% \multirow{2}{*}{\textbf{Batch Size}} & \textbf{1 Local Epoch} & \textbf{5 Local Epochs} & \textbf{10 Local Epochs} \\
% \cline{2-4}
% & Acc(\%) $\pm$ SD & Acc(\%) $\pm$ SD & Acc(\%) $\pm$ SD \\
% \hline
% 64 & 93.5 $\pm$ 2.22 & 93.9 $\pm$ 1.91 & 92.9 $\pm$ 1.81 \\
% \hline
% \end{tabular}
% \caption{TabLeak Attack Accuracy on Baseline}
% \label{tab:tableak_results}
% \end{table}

\begin{table}[h]
\renewcommand{\arraystretch}{1.5}
\centering
\small
\begin{tabular}{|c|c|c|}
\hline
\multirow{2}{*}{\textbf{Batch Size}} & \textbf{Baseline (FedAvg)} & \textbf{DPxFin} \\
\cline{2-3}
& Acc(\%) $\pm$ SD & Acc(\%) $\pm$ SD \\
\hline
64 & 92.9 $\pm$ 1.81 & 58.5 $\pm$ 2.2 \\
\hline
\end{tabular}
\vspace{2mm}
\caption{TabLeak Attack Results}
\label{tab:tableak_results}
\end{table}

% \begin{table}[!h]
% \renewcommand{\arraystretch}{1.2}
% \large
% \centering
% \begin{tabular}{|c|c|}
% \hline
% \textbf{Noise Scale} & \textbf{TabLeak Accuracy (\% ± SD)} \\
% \hline
% $0.2$ & 58.4 ± 2.4 \\
% \hline
% $0.5$ & 58.6 ± 2.6 \\
% \hline
% $1.0$ & 58.5 ± 2.2 \\
% \hline
% \end{tabular}
% \caption{DPxFin's Resilience to TabLeak Attack}
% \label{tab:noise_scale_results}
% \end{table}

%-- Line Plots

%----- Classification Report Table -----%
Table~\ref{tab:report-iid-noniid} reports the performance of the best-performing global model, evaluated on an independent test dataset following the completion of training, under both IID and non-IID settings. A maximum improvement of around 2\% was achieved in Non-IID setting. The evaluation metrics—Accuracy, F1 Score, Precision, and Recall—demonstrate consistently strong results across all criteria. These findings validate the robustness and generalization capability of our model, confirming its effectiveness in diverse data distribution scenarios.

\begin{table*}[!h]
\renewcommand{\arraystretch}{1.2}
\setlength{\tabcolsep}{4pt}
\centering
\begin{tabular}{|p{1.1cm}|c|c|c|c|c|c|c|c|c|c|c|}
\hline
\textbf{Setting} & \textbf{Clients} & \textbf{Method} & 
\multicolumn{4}{c|}{\cellcolor{gray!10}\textbf{No Privacy}} & 
\multicolumn{4}{c|}{\cellcolor{gray!45}\textbf{$\sigma = 1.0$}} \\
\cline{4-11}
 & & & 
\textbf{Accuracy} & \textbf{F1 Score} & \textbf{Precision} & \textbf{Recall} & 
\textbf{Accuracy} & \textbf{F1 Score} & \textbf{Precision} & \textbf{Recall} \\
\hline
\multirow{9}{*}{IID} 
 & \multirow{3}{*}{5} & FedAvg & 0.94 & 0.94 & 0.94 & 0.94 & -- & -- & -- & -- \\
 & & DP-FedAvg & -- & -- & -- & -- & 0.89 & 0.90 & 0.90 & 0.89 \\
 & & DPxFin & -- & -- & -- & -- & 0.91 & 0.91 & 0.91 & 0.91 \\
\cline{2-11}
 & \multirow{3}{*}{10} & FedAvg & 0.93 & 0.93 & 0.93 & 0.93 & -- & -- & -- & -- \\
 & & DP-FedAvg & -- & -- & -- & -- & 0.90 & 0.90 & 0.91 & 0.90 \\
 & & DPxFin & -- & -- & -- & -- & 0.91 & 0.91 & 0.91 & 0.91 \\
\cline{2-11}
 & \multirow{3}{*}{15} & FedAvg & 0.93 & 0.93 & 0.93 & 0.93 & -- & -- & -- & -- \\
 & & DP-FedAvg & -- & -- & -- & -- & 0.88 & 0.88 & 0.89 & 0.88 \\
 & & DPxFin & -- & -- & -- & -- & 0.89 & 0.89 & 0.89 & 0.89 \\
\hline

\multirow{9}{*}{Non-IID} 
 & \multirow{3}{*}{5} & FedAvg & 0.94 & 0.94 & 0.94 & 0.94 & -- & -- & -- & -- \\
 & & DP-FedAvg & -- & -- & -- & -- & 0.87 & 0.87 & 0.88 & 0.87 \\
 & & DPxFin & -- & -- & -- & -- & 0.89 & 0.89 & 0.90 & 0.89 \\
\cline{2-11}
 & \multirow{3}{*}{10} & FedAvg & 0.93 & 0.92 & 0.92 & 0.93 & -- & -- & -- & -- \\
 & & DP-FedAvg & -- & -- & -- & -- & 0.88 & 0.88 & 0.89 & 0.88 \\
 & & DPxFin & -- & -- & -- & -- & 0.90 & 0.90 & 0.90 & 0.90 \\
\cline{2-11}
 & \multirow{3}{*}{15} & FedAvg & 0.92 & 0.92 & 0.92 & 0.92 & -- & -- & -- & -- \\
 & & DP-FedAvg & -- & -- & -- & -- & 0.90 & 0.90 & 0.90 & 0.90 \\
 & & DPxFin & -- & -- & -- & -- & 0.91 & 0.91 & 0.91 & 0.91 \\
\hline
\end{tabular}
\caption{Classification Report of Best Performing Global Model under IID and Non-IID Settings}
\label{tab:report-iid-noniid}
\end{table*}

\newpage
\section{Conclusions}
In this work, we introduced \textbf{DPxFin}, a novel reputation-guided adaptive differential privacy framework specifically designed for federated learning in anti-money laundering detection systems. The proposed method addresses the critical challenge of balancing privacy preservation with model utility in the context of anti-money laundering (AML) detection, where sensitive financial data must be protected while maintaining effective collaborative learning across financial institutions. The reputation-based noise allocation approach is especially effective, and it enables high-performing clients to make a greater contribution to the global model to implement stricter privacy protection on potentially unreliable participants. The security analysis through TabLeak attack \cite{vero2023tableak} provides compelling evidence of our framework's robustness against sophisticated privacy attacks, with consistently low reconstruction accuracy (approximately 58.5\%), confirming the effectiveness of our privacy-preserving mechanism. The reputation-weighted aggregation mechanism does not only increase privacy, but also results in the improved robustness of the model due to decreased impact of possibly malicious or low-quality updates. Our work has broader implication than just detecting money laundering, as the implications are in other privacy sensitive collaborative learning settings laying the basis to future work in developing more advanced reputation-based schemes, and extending to other subjects of high stakes where privacy-preserving collaborative learning is critical.

%------------------------------------------------------%
%-------------------- Acknowledgments -----------------%
%------------------------------------------------------%
\section*{Acknowledgments}
This Research is supported by the RIE2025 Industry Alignment Fund – Industry Collaboration Project (IAF-ICP) (Award No: I2301E0020) and Japan-Singapore Joint Call: Japan Science and Technology Agency (JST) and Agency for Science, Technology and Research (A*STAR) 2024 (Award No: R24I6IR141), administered by A*STAR.

\bibliographystyle{plain}
\bibliography{references}

\end{document}